\pdfoutput=1

\documentclass[11pt]{article}

\usepackage[]{authblk}
\usepackage{mdframed}
\usepackage{setspace}
\usepackage{multirow}
\usepackage[table,xcdraw]{xcolor}

\makeatletter
\renewcommand\AB@affilsepx{, \protect\Affilfont}
\makeatother

\usepackage[]{ACL2023}
\let\oldfootnotetext\footnotetext
\renewcommand{\footnotetext}[1]{%
  \begingroup%
  \renewcommand{\thefootnote}{\ensuremath{*}}%
  \oldfootnotetext{#1}%
  \endgroup%
}

\usepackage{times}
\usepackage{latexsym}
\usepackage{graphicx}
\usepackage{tabu}
\usepackage{multirow}
\usepackage{makecell} 

\usepackage[T1]{fontenc}

\usepackage[utf8]{inputenc}

\usepackage{microtype}

\usepackage{inconsolata}
\usepackage{csquotes}

\usepackage{todonotes}

\title{HeGeL: A Novel Dataset for Geo-Location from Hebrew Text}

\author[,a]{Tzuf Paz-Argaman*}
\author[,b]{Tal Bauman*}
\author[a]{Itai Mondshine}
\author[c]{\authorcr Itzhak Omer}
\author[b]{Sagi Dalyot} 
\author[a]{Reut Tsarfaty}

\affil[a]{Bar-Ilan University, Israel}
\affil[b]{The Technion, Israel}
\affil[c]{Tel Aviv University, Israel}

\affil[ ]{\authorcr \tt \{tzuf.paz-argaman, mondshi1, reut.tsarfaty\}@biu.ac.il}

\affil[ ]{\authorcr \tt \ talbauman@campus.technion.ac.il}

\affil[ ]{ \tt \ omery@tauex.tau.ac.il}
\affil[ ]{\authorcr  \tt \ dalyot@technion.ac.il}

\begin{document}
\maketitle
\begin{abstract}
The task of textual geolocation — retrieving the coordinates of a place based on a free-form language description — calls for not only grounding but also natural language understanding and geospatial reasoning.
 Even though there are quite a few datasets in English used for geolocation, they are currently based on open-source data (Wikipedia and Twitter), where the location of the described place is mostly implicit, such that the location retrieval resolution is limited. Furthermore, there are no datasets available for addressing the problem of textual geolocation in morphologically rich and resource-poor languages, such as Hebrew. In this paper, we present the Hebrew Geo-Location (HeGeL) corpus, designed to collect literal place descriptions and analyze lingual geospatial reasoning. We crowdsourced 5,649 literal Hebrew place descriptions of various place types in three cities in Israel.
 Qualitative and empirical analysis show that the data exhibits abundant use of geospatial reasoning and requires a novel environmental representation.\footnotetext{Equal contribution.}\footnote{For data and code see https://github.com/OnlpLab/HeGeL
}
\end{abstract}

\section{Introduction and Background}

 \begin{figure*}[th]


  \centering
\scalebox{0.99}{
            \includegraphics[width=1 \textwidth]{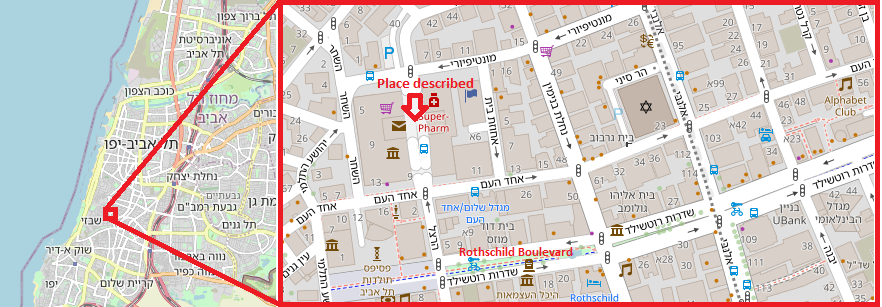}}
            
            \vspace*{-1mm}
                 {\footnotesize{

                 \begin{spacing}{0.5}
                
                 \begin{flushleft} 
 \begin{mdframed}
 
 \textbf{Place Description:}       
 The place is located near the Rothschild complex -- at the end of Rothschild Street, as you go towards the sea, take a right for about three streets and then you will see the tower high above you.
 \end{mdframed}
 \end{flushleft} 
\end{spacing} }}

        \caption
        {A description example from HeGeL translated from Hebrew.
        } 

        \label{fig:hegel_example}
    \end{figure*}

Textual Geolocation Identification, a crucial component of Geographic Information Retrieval (GIR), is the task of resolving the location, i.e., coordinates of a place, based on the reference to it  in a text. It requires a combination of language and environmental knowledge. 
On top of the usual non-spatial linguistic challenges in Natural Language Understanding (NLU), such as named entity recognition (NER), anaphora resolution, bridging anaphora, etc., the textual geolocation task presents geospatial challenges
 that require multimodal processing and grounding \cite{ji2022abstract, antol2015vqa, misra2017mapping, qi2020reverie, paz2020zest}. 

Proper names, such as \enquote*{Rabin Square}, also known as {\em named entities} in Natural Language Procesing (NLP), and as {\em rigid designators} in formal semantics \cite{kripke1972naming}, can be easily grounded based on a Gazetteer or a simple map. 
However, geolocating linguistic terms that involve spatial expressions without the explicit mention of a proper name still present an open challenge. This interpretation challenge includes the understanding and resolution of (at least): (i) definite descriptions, such as \enquote*{the school} (ii) geospatial terms, such as cardinal directions; \enquote*{east of};  and (iii) geospatial numerical reasoning; \enquote*{two buildings away from the pharmacy}. 
To address these and other challenges, we need to both ground entity mentions to their corresponding physical entities in the environment, and to reason about geospatial relations expressed between entities —  these two processes being closely intertwined. 

To do so, we need a corpus for the geolocation task that maps rich geospatial place descriptions to their corresponding location coordinates. However, current corpora for  geolocation  are based on naturally-occurring open-source resources, such as Wikipedia articles \cite{eisenstein2010latent, wing2011simple, han2012geolocation, wing2014hierarchical, wallgrun2018geocorpora}, which are not spatially oriented, i.e., the description of locations is implicit or absent in the corresponding text. Subsequently, the accuracy of retrieval is fairly low (around 100 km).

Furthermore, all geolocation datasets previously studied in NLP are in English, with a dearth of corpora for low-resource languages, in particular, for morphologically rich languages, such as Hebrew.  
To understand the geolocation challenges and build models that do various spatial reasoning tasks, English cannot be our sole focus \cite{baldridge2018points}. Hebrew,
a Semitic morphologically rich language is
notoriously difficult to parse \cite{tsarfaty2020spmrl, tsarfaty2019s}. Moreover, resources that are available for Hebrew NLP research focus on traditional tasks, such as Part-of-speech (POS) tagging, syntactic parsing, etc; and lack corpora for understanding and reasoning in real-world situations.

In this work we present HeGeL, a novel dataset for \textbf{He}brew \textbf{Ge}o-\textbf{L}ocation, the first ever Hebrew NLU benchmark involving both grounding and geospatial reasoning.
To create HeGeL, we crowdsourced 5,649 geospatially-oriented Hebrew place descriptions of various place types from three cities in Israel.
We designed our task based on a realistic scenario of human place description, relying on people’s memory of the world, rather than, e.g., using a map \citep{anderson1991hcrc, paz2019run}.
Crucially, relying on environmental cognition results in various levels of geospatial knowledge \cite{siegel1975development} that are manifested in the descriptions and the geospatial reasoning that is required to resolve their location \cite{hayward1995spatial}.  
To avoid the much simpler task of grounding proper named entities, 
we explicitly restricted the use of proper names in the  description of the place and adjacent landmarks.

%
%
Unlike the text-based navigation task \cite{macmahon2006walk, chen2019touchdown, ku2020room, de2018talk, thomason2020vision}, which requires representing an agent's current perspective, reflecting its route knowledge, we show that the HeGeL task requires a full-environment representation, thus, capturing complex geospatial relations among multiple physical entities. Through a thorough linguistic and empirical analysis, we demonstrate the characteristics and challenges associated with Hebrew place descriptions, showing that HeGeL serves both as a challenging NLU benchmark and as a corpus for geospatial cognition research. 

\section{The HeGeL Task and Dataset}

This work addresses the task of geolocating places on a map  based on natural language (NL) geospatial descriptions that are given in a colloquial language and based on participants' memory of the environment (i.e., cognitive map). The input to the HeGeL task is as follows: (i) an NL place description of the whereabouts of the place, and (ii) a map with rich details of the environment (e.g.,  physical entities names, geospatial relations, and attributes). The output is a pair of coordinates (x,y) specifying the physical location of the place described in the text. Figure \ref{fig:hegel_example} shows an example of a place description from HeGeL translated from Hebrew.

To simplify the crowdsourcing task and encourage participants’ engagement, we frame the data crowdsourcing process as the well-known game, the {\em treasure hunt} task \cite{kniestedt2022re}, in which the {\em instructor-participant} is required to describe in writing the location of the treasure, a known place in the city, to a different {\em follower-participant}   who then needs to locate it on a map. Thus, the online assignment is divided into two tasks: the instructor's \textit{writing of place descriptions} and the follower's \textit{validation}. To avoid preconceived notions as to the \enquote*{correct} way to describe a place, we first presented the participants with the task of writing a place description, and once completed, the validation task was given.\footnote{Appendix \ref{appendix:data_collection} includes additional data collection details. Appendix \ref{appendix:ui} presents a display of the online assignment's UI translated from Hebrew to English.}



\noindent
We hereby provide the details of the two   UI tasks:
\paragraph{\em (i) Task 1. Writing a place description}
In this task we requested participants to describe in a free-form text the location of a place known to them, to a third party who might not be familiar with the whereabouts of that place. To collect place descriptions based solely on people’s memory, we did not visualize the area of the place, e.g., on a map. Instead, we ensured that the participants are well familiarized with the place by asking them to state how familiar they are with the place on a scale of 1-5. If this score was 1 or 2, we presented the participant with a different place to describe. 
To ensure diverse human-generated textual descriptions, places were chosen based on their type, position/location in the city (places were spread across  the city), geometry, size, and context. To avoid the use of proper names, we developed a rule-based methodology to make sure that the explicit name of the goal (place) or of the nearby landmarks (< 100 meters) will not appear explicitly in the description. The original description was saved, and the participants were asked to input another description without the above names.

\paragraph{\em (i) Task 2. Place description validation}
To verify that a person who reads the text description will understand where the treasure is hidden, i.e., geolocate the place, we developed a map-based retrieval task. The participant in the follower role was asked to read the crowdsourced textual description and mark its location on the map, i.e., where the treasure is hidden. For marking the location, we implemented an interactive online map based on \href{http://www.openstreetmap.org}{OpenStreetMap (OSM)},\footnote{OSM is a free, editable, map of the whole world, that was built by volunteers, with millions of users constantly adding informative tags to the map.} which allows the participants to move and zoom-in to precisely pin the described place on the map. The map supports the cognitive process needed to ground mentioned  entities to physical entities, reason about the geospatial relations, and locate the described place. 
To familiarize participants with the interactive map tool and task, they had to first pass a simple map marking test, and only then they could start task 2 of  reading place descriptions (given by other participants), marking place locations on the map, and rate the clarity of the textual description on a scale of 1-5.

\begin{table*}[t]
\scalebox{0.82}{
\begin{tabular}{lllll}
\hline
\textbf{Phenomenon}                         &           & \multicolumn{1}{c}{\textbf{\textit{c}}} & \multicolumn{1}{c}{\textit{$\mu$}} & \multicolumn{1}{c}{\textbf{Example from HeGeL (translated into English)}}         \\ \hline
\multirow{5}{*}{\begin{tabular}[c]{@{}l@{}}Type of \\ elements in a city \\ \cite{lynch1960image} \end{tabular}}  & Edge      & 36\%                              & 0.6                            & \enquote{when reaching Yafo, one should go toward the \textbf{sea}…}               \\
                                            & Node      & 40\%                             & 0.44                           & \enquote{…a few minutes walk from the \textbf{HaShaon square}…}                    \\
                                            & Landmark  & 60\%                             & 1.08                           & \enquote{…near \textbf{Levinski market}}                                           \\
                                            & District  & 36\%                              & 0.4                            & \enquote{\textbf{South part of the city} next to…}                                 \\
                                            & Path      & 68\%                             & 0.76                           & \enquote{On \textbf{Carlebach street}…}                                           \\
\multirow{3}{*}{\begin{tabular}[c]{@{}l@{}}Spatial knowledge \\ \cite{siegel1975development}  \end{tabular}}           & Landmarks & 32\%                              & n/a                              & \enquote{Next to the \textbf{sea} in Tel Aviv-Yafo}                                \\
                                            & Route     & 20\%                              & n/a                              & \enquote{\textbf{Passing Azrieli on Menachem Begin and then turn right…}} \\
                                            & Survey    & 48\%                             & n/a                              & \enquote{\textbf{South part of the city near Levinski market}}            \\
\multicolumn{2}{l}{Reference to unique entity}          & 100\%                             & 2.32                           & \enquote{…in the middle of \textbf{Dizengoff street}}                              \\
\multicolumn{2}{l}{Cardinal direction}                  & 44\%                             & 0.76                           & \enquote{\textbf{South} of Sharona…}                                               \\
\multicolumn{2}{l}{Coreference}                         & 16\%                              & 0.16                           & \enquote{…continue a bit west and \textbf{it}…}                                    \\ \hline

\end{tabular}
}
\caption{Linguistic qualitative analysis of 25 randomly sampled descriptions in HeGeL. \textit{c} is the percentage of descriptions containing at least one example of the phenomenon, and \textit{$\mu$}  is the mean number of times the phenomenon appears in each description.}
\label{tab:qualitive}
\end{table*}

\paragraph{Target Selection and Retrieval Errors} The treasure-hunt task we devised included 167 places in the three largest cities in Israel: Tel Aviv, Haifa, and Jerusalem. These three cities are differently shaped, and show different physical, morphological and topographic features, which potentially affect the legibility and imageability of urban components, and therefore also on place descriptions. These differences can be expressed in the use of various physical features and prepositions, e.g., frequent use of the physical object \enquote*{landmark} and the prepositions \enquote*{above} or \enquote*{below} in hilly terrains that characterize Haifa and Jerusalem.

To assess the quality and interpretability of the place descriptions,
we calculate the shortest Euclidean distance between the coordinates of the goal's (physical element) shape (polygon, line or point), and the location marked by the 'follower' on the map (task 2); we term this distance as {\em retrieval error}. To determine the agreement rate among human participants, each textual place description is validated by at least two participants. To ensure that we work with descriptions that can be geolocated, we set a hard distance threshold  of 300 meters, based on analysis of the descriptions' clarity score that we had conducted on a prior (held-out)  development corpus we collected for the task.    

\section{Data Statistics and Analysis}

The resulting HeGeL dataset contains 5,649 validated descriptions paired with their coordinates on a map. The locations are divided among three cities: 2,142 in Tel Aviv, 1,442 in Haifa, and 2,065 in Jerusalem. 1,833 participants completed the writing task, inserting in total 10,946 place descriptions, and 2,050 participants completed 12,655 validation tasks. The dataset is balanced, with about 33 descriptions per place.

\begin{figure}[t]
\centering
\scalebox{0.49}{
\includegraphics[width=\textwidth]{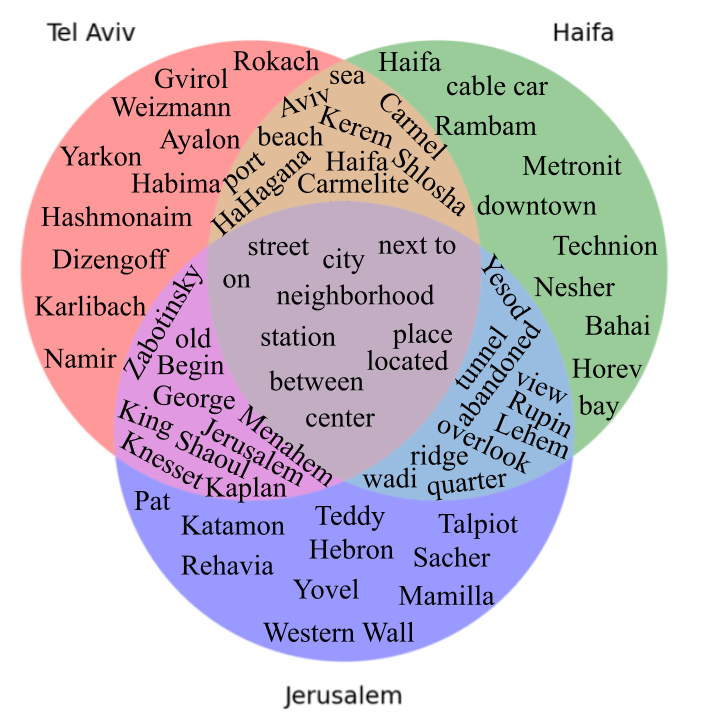}}
 \caption{A Venn diagram showing the top 10 words (translated from Hebrew) used in each city-set relation. 
 } 
\label{fig:venn}%
\end{figure}

Figure \ref{fig:venn} shows a Venn diagram representing the relation of the three sets of city-based vocabularies (formed from unique lemmas produced by \citet{more2019joint} lemmatization tool). The intersection of the three cities contains only 15.07\% of the entire vocabulary (the union of the three cities’ vocabularies). 
The shared language is not focused on city-specific terms, 
such as \enquote*{Knesset}. Instead, it includes rich spatial terms, such as \enquote*{between}, modified prepositions such as \enquote*{next to}, and non-definite entities, such as \enquote*{street}.
From the Venn diagram we also conclude that almost half of the lemmas of the three vocabularies, corresponding to the three cities, contain city-specific lemmas: 48.6\%, 40.65\%, and 49.3\% for Tel Aviv, Haifa, and Jerusalem, respectively. As such, HeGeL enables a city-split setup, training on one city and testing on a different unseen city, where city-reserved named entities present an out-of-vocabulary (OOV) challenge for models trained on another city.

\begin{table}[t]
\scalebox{0.66}{
\begin{tabular}{lcc }
\Xhline{6\arrayrulewidth}
\textbf{Feature}        & \textbf{Avg. per description} & \textbf{Unique in corpus} \\ \hline
{Number of lemmas}         &  12.93  & {6,663}                      \\
{Number of tokens}         & {11.50}                          & {9,207}                      \\
{Number of named entities} & {0.55}                           & {3,490}                      \\
{Number of prepositions}   & {2.39}                           & { 14,256}                     \\
{Number of verbs}          & {0.53}                           & {3,152}                      \\ \Xhline{6\arrayrulewidth}
\end{tabular}
}
\caption{Quantitative analysis of HeGeL.}
\label{tab:Quantitive}

\end{table}

\begin{table}[t]
\scalebox{0.53}{
\begin{tabular}{llcc}
\Xhline{6\arrayrulewidth}

\rowcolor[HTML]{FFFFFF} 
\multicolumn{2}{l}{\cellcolor[HTML]{FFFFFF}\textbf{Feature}}                                                                                                                                                                       & \textbf{p-value}                                               & \textbf{FDR corrected p-value}                                 \\ \hline
\rowcolor[HTML]{FFFFFF} 
\cellcolor[HTML]{FFFFFF}{\color[HTML]{202124} }                                                                                                                         & {\color[HTML]{202124} Number of Words}                   & {\color[HTML]{202124} 0.8306}                                  & {\color[HTML]{202124} 0.8306}                                  \\
\rowcolor[HTML]{FFFFFF} 
\cellcolor[HTML]{FFFFFF}{\color[HTML]{202124} }                                                                                                                         & {\color[HTML]{202124} \textbf{Number of named entities}} & {\color[HTML]{202124} \textbf{0.0000}}                         & {\color[HTML]{202124} \textbf{0.0000}}                         \\
\rowcolor[HTML]{FFFFFF} 
\cellcolor[HTML]{FFFFFF}{\color[HTML]{202124} }                                                                                                                         & {\color[HTML]{202124} \textbf{Number of prepositions}}   & {\color[HTML]{202124} \textbf{0.0145}}                         & {\color[HTML]{202124} \textbf{0.0217}}                         \\
\rowcolor[HTML]{FFFFFF} 
\multirow{-4}{*}{\cellcolor[HTML]{FFFFFF}{\color[HTML]{202124} \textbf{\begin{tabular}[c]{@{}l@{}}Task 1: \\ Place description \\ (linguistic features)\end{tabular}}}} & {\color[HTML]{202124} Number of verbs}                   & {\color[HTML]{202124} 0.3400}                                  & {\color[HTML]{202124} 0.4080}                                  \\ \hline
\rowcolor[HTML]{FFFFFF} 
\cellcolor[HTML]{FFFFFF}{\color[HTML]{202124} }                                                                                                                         & {\color[HTML]{202124} \textbf{Retrieval error}}          & \cellcolor[HTML]{FFFFFF}{\color[HTML]{202124} \textbf{0.0000}} & \cellcolor[HTML]{FFFFFF}{\color[HTML]{202124} \textbf{0.0000}} \\
\rowcolor[HTML]{FFFFFF} 
\multirow{-2}{*}{\cellcolor[HTML]{FFFFFF}{\color[HTML]{202124} \textbf{\begin{tabular}[c]{@{}l@{}}Task 2: \\ Human verification\end{tabular}}}}                         & {\color[HTML]{202124} \textbf{Clearness score}}          & \cellcolor[HTML]{FFFFFF}{\color[HTML]{202124} \textbf{0.0000}} & \cellcolor[HTML]{FFFFFF}{\color[HTML]{202124} \textbf{0.0000}}

\\ \Xhline{6\arrayrulewidth}
\end{tabular}
}
\caption{Correlations between place types and linguistic and features.}
\label{tab:anova}

\end{table}

Table \ref{tab:qualitive} shows an analysis of the linguistic phenomena manifested in the HeGeL dataset, 
demonstrating the spatial knowledge and reasoning skills required for solving the HeGeL task. We analyzed the frequency of the five types of elements in a city defined by \citet{lynch1960image}, along with the three types of spatial knowledge defined in \citet{siegel1975development}, and other spatial properties. 
The frequent use of cardinal directions, as well as the use of survey knowledge, suggests that any NLP model built to deal with the HeGeL task should not only represent a local view of the goal, or possible routes, but also take into consideration the full region, and mimic people’s map-like view of the environment. Therefore, unlike navigation tasks where only the agent's current perspective is represented in the model, this task requires full representation of the environment. \par
We further perform a quantitative analysis of word tokens and lemmas that appear in HeGeL, depicted in Table \ref{tab:Quantitive}. Overall, the HeGeL dataset contains a large vocabulary of 9,207 unique tokens and 6,663 unique lemmas. There are mentions of physical entities, but as we limited the mentions of named-entities of the described place and landmarks adjacent to it; these are relatively rare, and are mostly references to prominent city landmarks. Also, as most place descriptions are not route-based descriptions, there are only few verbs used in the descriptions. Prepositions, on the other hand, are abundant.

In Table \ref{tab:anova}, using a one-way analysis of variance (ANOVA) test, we found a significantly (p<0.05) different distribution between place type descriptions and the following features: number of named entities, number of verbs, human verification retrieval error, and clarity score.

\section{Experiments}

We create a zero-shot (ZS) city-based split, such that we train on one city and test on another. The train, development, and test sets correspond to the descriptions collected in Tel Aviv, Haifa, and Jerusalem, respectively. 
We evaluate different baseline models for the geolocation task on the HeGeL dataset. We use three evaluation metrics based on retrieval error: mean, median, and task completion (TC) accuracy -- the percentage of place descriptions located within the 300 meters threshold. We provide three baselines for the HeGeL task. 

We first assess a brute-force NER approach; i.e., we test whether recognizing named entities in the text and retrieving their corresponding coordinates is sufficient for solving the HeGeL task of geolocation. 
To this end, we used Google Maps API and produced two baseline models: (i) Google Maps API Query — we queried the API with the full raw text descriptions as input, with no prepossessing; and (ii) Oracle NER — we queried all 1-5 n-grams against Google Maps API and retrieved the closest geolocation to the goal.

In our second approach, we employ a dual-encoder model. One encoder encodes the text using a Hebrew Monolingual pre-trained encoder, AlephBERT \cite{seker2022alephbert}, which produces a 768-dimension vector representation of the text. The other encoder processes the environment, which is represented as a graph based on OSM data.
Each point of interest in the graph is connected to an S2Cell\footnote{S2Cells are based on S2-geometry (https://s2geometry.io/), a hierarchical discretization of the Earth’s surface  \citep{hilbert1935stetige}.}, which contains its geometry and is based on S2-geometry. These S2Cells are encoded using a random-walk algorithm to produce a 64-dimensional vector for each cell. These vectors are then passed through a linear layer to produce 768-dimensional vectors.  We calculate the cosine similarity score between the text and environment vectors and use it to align the respective representations  via maximization of the cosine similarity score with a cross-entropy loss over the scores.

Performing an ANOVA test, we found a significantly (p<0.05) different distribution between place type descriptions and the retrieval error of the Oracle NER. The mean retrieval error of the Path and Node place types were the lowest in both human verification and Oracle NER. This suggests that both of these place types are easier for humans to geolocate. 


\begin{table}[t]
\centering
\scalebox{0.63}{
\begin{tabular}{lcccc}
\Xhline{6\arrayrulewidth}
\textbf{Split}      & \textbf{Model}                & \textbf{Mean} & \textbf{Median} & \textbf{TC}  \\ \hline
\multirow{3}{*}{ZS} & Google Maps API Query         & 2,811         & 849             & 27.66      \\
                    & Oracle NER*                   & 2,373         & 496             & 37.79    \\
                    & HUMAN                         & 553           & 151             & 70.81**      \\ \hline
ZS                  & \multirow{3}{*}{Dual-encoder} & 2,727\small(1684)   & 2,612\small(1930)     & 2.37\small(1.5)  \\
FS 20\%       &                               & 1717\small(35)      & 1583\small(49)        & 3.43\small(0.09) \\
FS 80\%       &                               & 983\small(23)       & 632\small(13)         & 15.7\small(0.38) \\ \Xhline{6\arrayrulewidth}
\end{tabular}
}
\caption{Baseline results over the zero-shot (ZS) city-split, and few-shot (FS) split of different sizes: 20\% and 80\% of the samples in the test-region. For the Dual-encoder we report the mean over three random initialization and the standard-deviation (std) is in brackets. *Oracle NER 
is a skyline model based on a NER approach. **The human agreement rate. }
\label{tab:results}

\end{table}



The results in Table \ref{tab:results} show that our task is not solvable with adequate resolution by the Google Maps API. The human performance provides an upper bound for the HeGeL task performance, while the simple Google Maps API Query provides a lower bound. The Google API model's low performance suggests that NER and the Gazetteer-based methods
in and of themselves are insufficient to handle the HeGeL task successfully, and that geospatial reasoning is necessary. The Dual-encoder's
low performance on the ZS split suggests that OOV is a major challenge. The few-shot (FS) split shows an improvement of the model after fine-tuning on additional samples from the test-region (FS 20\% and 80\%). This 
 suggests that a possible solution for the city-split setup might be data-augmentation via generating grounded descriptions for the tested region -- an approach we reserve for future research.  

\section{Conclusion}

The contribution of this paper is threefold.
First, we present the first geolocation benchmark with Hebrew place descriptions. Second, to the best of our knowledge, this is the only  {\em crowdsourced} geolocation dataset, thus, eliciting explicit geospatial descriptions, allowing for better retrieval resolution. Finally,  our analysis shows that the dataset presents complex spatial reasoning challenges which require novel environmental model representation.

\section*{Limitations}
While we aim for our HeGeL crowdsourcing methodology to be applicable to other languages, and in particular low-resource languages, the UI design and our analyses require knowledge of the intended language, as well as familiarity with the regions where it is spoken. Moreover, as our methodology relies on people's familiarity with the places, it limits the cities chosen for the task and the participants that could take part, restricting the demographics of the participants accordingly. In addition, relying on people's memory of the environment causes many of the descriptions to be too vague for humans to geolocate, thus, many of the descriptions were disqualified during the validation process as they could not have been resolved. The relatively low percentage of place descriptions that were successfully validated, raises the costs of collecting such a dataset.

\section*{Acknowledgements}
This research is funded by a grant from
the European Research Council, ERC-StG grant
number 677352, and a grant by the Israeli Ministry of Science and Technology (MOST), grant number 3-17992, for which we are grateful.

\bibliography{anthology}
\bibliographystyle{acl_natbib}

\appendix
\section{Data Collection Details}
\label{appendix:data_collection}
We used the services of an Israeli surveying company to distribute the assignment to native Hebrew-speakers participants in Israel only. The survey company was charged with distributing the assignments to a balanced set of participants in terms of their demographic and geographic characteristics (e.g., an equal number of males and females). All participants were given full payment, non-respective of whether they correctly completed the task. 

The first page the participants viewed contains a disclosure about the assignments being part of academic research and the purpose of the assignments. The assignment protocol was approved by a behavioral review board. 
This approval was also presented to the participants on the initial screen. Also, the participants were required to read an informed consent form and sign an agreement box. 





\section{Participant Interface}
\label{appendix:ui}

\begin{figure}[h]
\centering
\scalebox{0.46}{
\includegraphics[width=\textwidth]{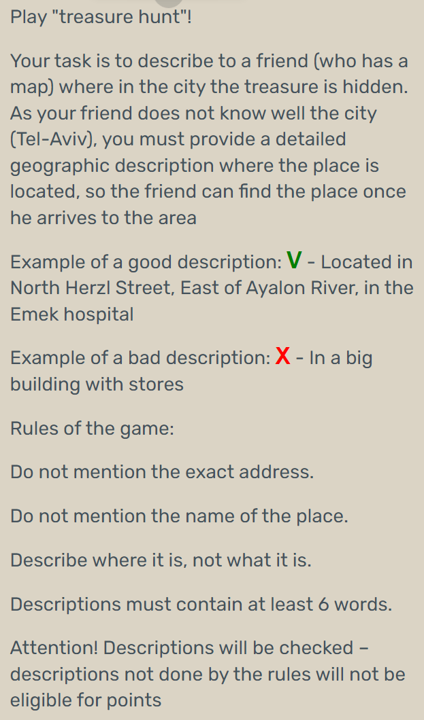}}
 \caption{Participant Interface translated from Hebrew: instructions for the writing task.
 } 
\label{fig:girit2}%
\end{figure}
\begin{figure}[b]
\centering
\scalebox{0.49}{
\includegraphics[width=\textwidth]{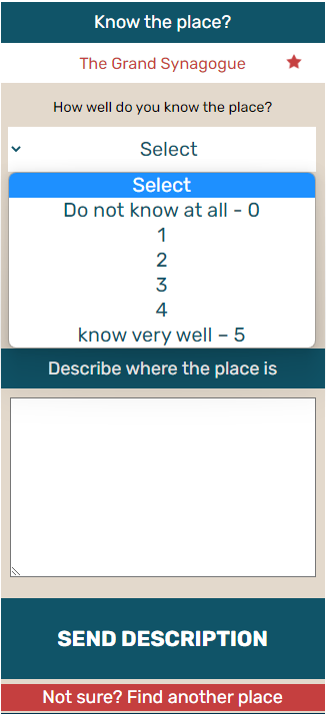}}
 \caption{Participant Interface translated from Hebrew: the writing task (task 1).
 } 
\label{fig:girit3}%
\end{figure}
\begin{figure}[b]
\centering
\scalebox{0.465}{
\includegraphics[width=\textwidth]{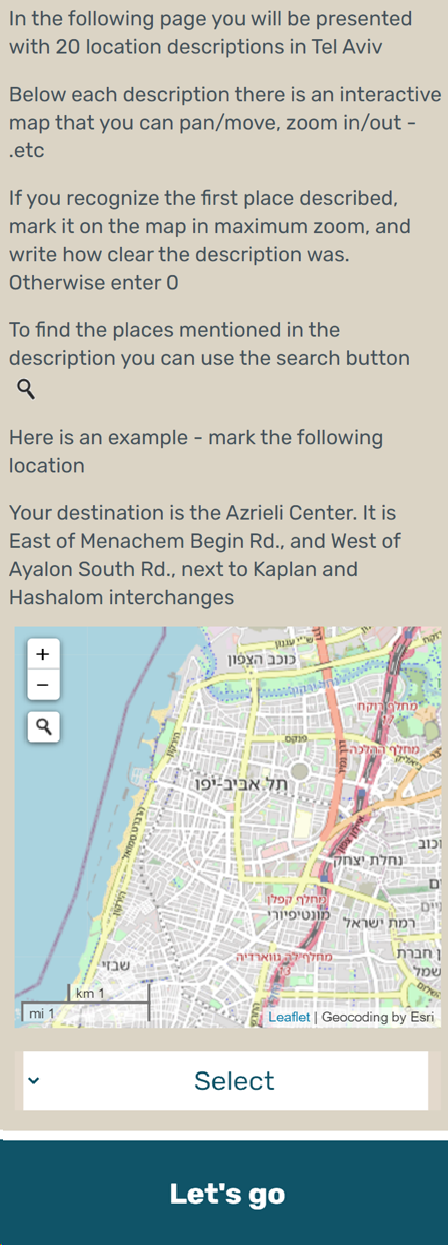}}
 \caption{Participant Interface translated from Hebrew: the validation task (task 2).
 } 
\label{fig:girit4}%
\end{figure}

The tasks are performed via an online assignment application, depicted in Figures \ref{fig:girit2}-\ref{fig:girit4}.


\section{Experimental Setup Details}

The cross-entropy loss function was optimized with Adam optimizer \citep{Adam}.
The hyperparameter tuning is based on the average results run with three different seeds.
The Learning rate was searched in [1e-5, 1e-4,  1e-3] and a 1e-5 was chosen.
The S2-cell level was searched in [13, 15, 17] and 13 was chosen.
Number-of-epochs for early stopping was based on their average learning curve.


\end{document}